\pdfoutput=1

\documentclass[11pt]{article}

\usepackage{ACL2023}

\usepackage{times} 
\usepackage{latexsym}
\usepackage{hyperref}
\usepackage{graphicx}
\usepackage{tabularx}
\usepackage{booktabs}
\usepackage{caption}
\usepackage{subcaption}
\usepackage{calc}
\usepackage{enumitem}
\usepackage{amsfonts}
\usepackage{amssymb}
\usepackage{xcolor}
\usepackage{multirow}
\usepackage{amsmath}
\usepackage{xspace}
\usepackage{todonotes}
\usepackage{soul}
\usepackage{arydshln}
\usepackage{pifont}
\usepackage{wrapfig}
\usepackage{booktabs}
\newcommand{\model}{\texttt{QUARC}\xspace}
\newcommand{\Pa}{\texttt{CLIME}\xspace}
\newcommand{\Pb}{\texttt{COGENT}\xspace}
\newcommand{\Paa}{\texttt{ITEM}\xspace}
\newcommand{\Pab}{\texttt{QUINCE}\xspace}
\newcommand{\Pba}{\texttt{TREAD}\xspace}
\newcommand{\fusion}{\texttt{PerFuMe}\xspace}
\newcommand{\semy}{z_i^s}
\newcommand{\syny}{z_i^f}
\newcommand{\cvec}{e_i^f}
\newcommand{\ivec}{\tilde{e}_i^f}
\newcommand{\semx}{\hat{x}_i^s}
\newcommand{\mapx}{\hat{z}_i^s}
\newcommand{\trgx}{\hat{t_i}}

\newcommand*{\Comb}[2]{{}^{#1}C_{#2}}

\newcommand{\R}[2]{#1 \in \mathbb{R}^{#2}}

\newlength\widest

\usepackage[T1]{fontenc}

\usepackage[utf8]{inputenc}

\usepackage{microtype}

\newcommand{\dataset}{\texttt{IntentCONAN}\xspace}

\title{\textit{Counterspeeches up my sleeve!} Intent Distribution Learning and Persistent Fusion for Intent-Conditioned Counterspeech Generation}

\author{Rishabh Gupta$^{1}$, Shaily Desai$^{1}$, Manvi Goel$^{1}$, Anil Bandhkavi$^2$, 
\\ \textbf{Tanmoy Chakraborty$^3$, \and Md Shad Akhtar$^{1}$} 
\\ $^1${IIIT Delhi, India}, $^2${Logically, U.K.}, $^3${IIT Delhi, India}
\\ \texttt{\{rishabh19089, shailyd, manvi19472, shad.akhtar\}@iiitd.ac.in}, \\  \texttt{anil@logically.ai}, \texttt{tanchak@iitd.ac.in}}

\begin{document}
\maketitle
\begin{abstract}
Counterspeech has been demonstrated to be an efficacious approach for combating hate speech. While various conventional and controlled approaches have been studied in recent years to generate counterspeech, a counterspeech with a certain intent may not be sufficient in every scenario. Due to the complex and multifaceted nature of hate speech, utilizing multiple forms of counter-narratives with varying intents may be advantageous in different circumstances. In this paper, we explore intent-conditioned counterspeech generation. At first, we develop \dataset, a diversified intent-specific counterspeech dataset with 6831 counterspeeches conditioned on five intents, i.e., \textit{informative}, \textit{denouncing}, \textit{question}, \textit{positive}, and \textit{humour}. Subsequently, we propose \model, a two-stage framework for intent-conditioned counterspeech generation. \model leverages vector-quantized representations learned for each intent category along with \fusion, a novel fusion module to incorporate intent-specific information into the model. Our evaluation demonstrates that \model outperforms several baselines by an average of \textasciitilde$10\%$ across evaluation metrics. An extensive human evaluation supplements our hypothesis of better and more appropriate responses than comparative systems.   

\textcolor{red}{\small \textit{Warning: This work contains offensive and hateful text that some might find upsetting. It does not represent the views of the authors.}}
\end{abstract}

\section{Introduction}
\label{intro}

\begin{figure}[ht]
     \centering
         \centering
         \includegraphics[width=\linewidth]{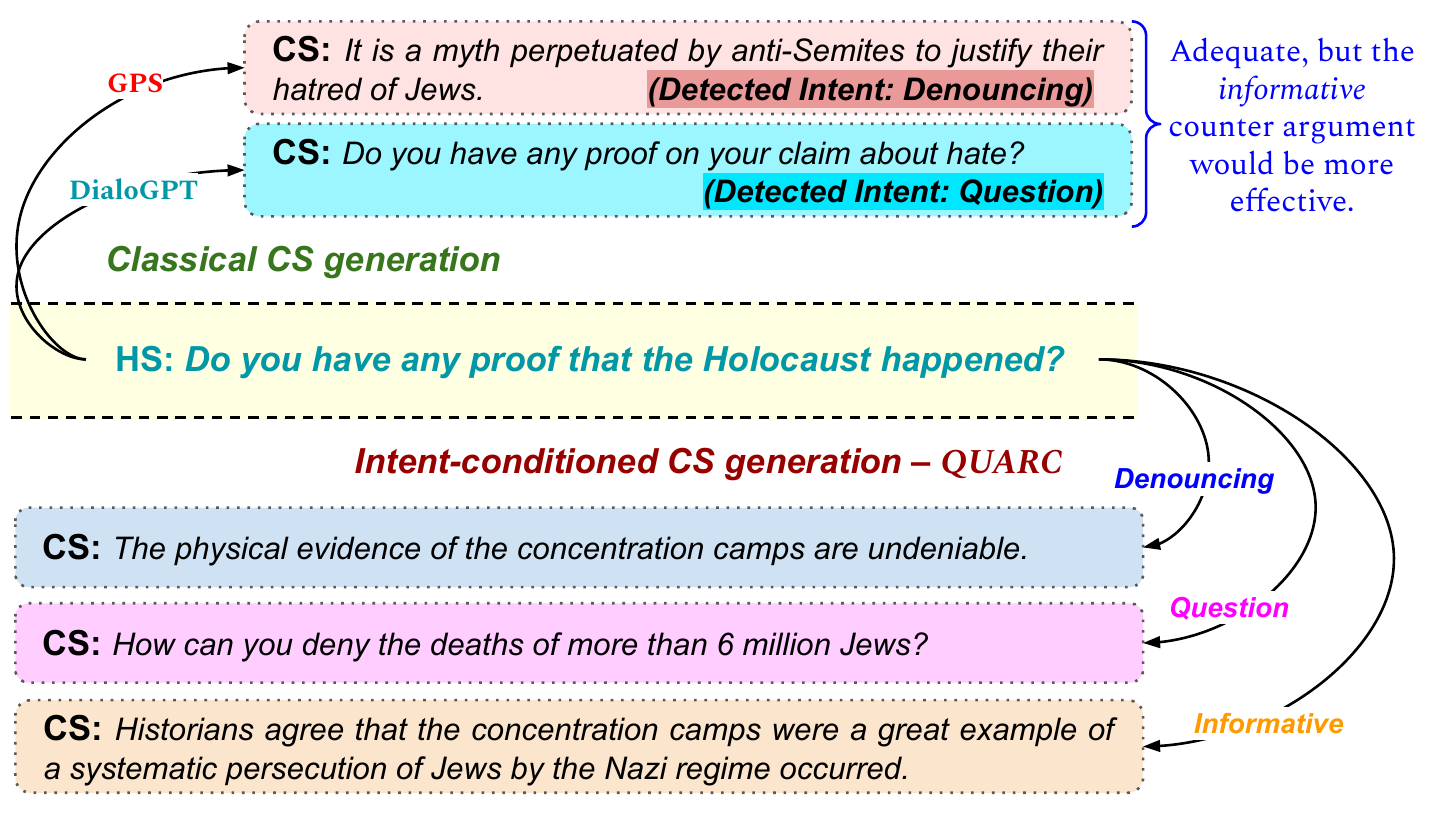}
     \caption{Outputs compared to pre-existing methods. These examples show different intents generated by different models. This raises the need for a system that, along with producing multiple counter-arguments, also ensures that the generated sentence is effective.}
     \label{fig:mot}
\end{figure}

The quantity and accessibility of information on the Internet are constantly growing in the 21st century. This has made it increasingly simpler for users on social media to post hateful or attacking speech, all while hiding behind the veil of anonymity \cite{10.1145/3078714.3078723}. Hate speech \cite{DBLP:conf/pakdd/AwalCLM21,ChakrabortyM22} is an offensive dialogue that uses stereotypes to communicate a hateful ideology, and it can target several protected qualities such as gender, religion, colour, and disability \cite{chetty2018hate}. This type of cyberhate could have long-term implications for both individuals and communities \cite{DBLP:conf/kdd/MasudBKA022}. Outlawing or regulating hate speech does not appear to be beneficial because it rarely improves the situation and may be interpreted as interfering with free speech \cite{10.1145/3134666}. Prohibiting hateful speech has also been demonstrated to have unexpected consequences, but more importantly, it introduces a curb to the opportunity to defend against potential harm with positive, unbiased, and informed statements that could incite change. The best strategy for fending off offensive online remarks is counterspeech \cite{wright-etal-2017-vectors,schieb2016governing}. Past initiatives such as WeCounterHate\footnote{\url{http://www.wecounterhate.com/}.} and GetTheTrollsOut\footnote{\url{https://getthetrollsout.org/}.}  have proven to make a difference; however, the sheer volume of online hate speech \cite{DBLP:journals/corr/abs-2103-11799} necessitates the development of a trustworthy and effective counterargument system.

\paragraph{Motivation:}
Every circumstance that necessitates counterspeech is distinct. Prior work \cite{zhu2021generate} in this domain is limited to generating one counterspeech instance for every hate speech. However, while appropriate, a single counterspeech style could fail to produce the desired effect on the attacker and bystanders alike. \newcite{mathew2019thou} showed that different victimized communities could be perceptible to different types of counterspeeches. The authors analyzed comments from YouTube and compared the popularity of various intents of counterspeeches for different affected communities like POC, LGBT+, and Jews. They concluded that most likes and replies were received by different kinds of counterspeech instances for different communities -- e.g., \textit{facts} and \textit{humor} in the case of LGBT+. These observations indicate that a counterspeech generation model would benefit from a diverse output pool, and generating appropriate counterspeeches for different scenarios would provide a better opportunity to educate the attacker and the general public. 
We support our argument with an example in Figure \ref{fig:mot}. For a given hate speech, we generate counterspeeches from Generate-Prune-Select (GPS) \cite{zhu2021generate} -- a popular counterspeech generation model, and fine-tuned DialoGPT \cite{zhang-etal-2020-dialogpt}. Though the counterspeeches with intents \textit{question} and \textit{denouncing}, respectively, are semantically appropriate and can be used as valid responses, we argue that the legitimacy of the evidence supporting the Holocaust would be best addressed by a \textit{factual/informative} counterspeech. To the best of our knowledge, {\em this paper presents the first successful pipeline for intent-controlled counterspeech generation}. 


\paragraph{Our Contribution:} We propose a novel task of \textbf{intent-specific counterspeech generation} that aims to generate a counterspeech for a given hate speech and a desired counterspeech intent. In total, we consider five counterspeech intents, namely -- \texttt{informative}, \texttt{question}, \texttt{denouncing}, \texttt{humor}, and \texttt{positive}. We curate \dataset, an \textit{intent-specific counterspeech generation dataset} consisting of $6,831$ counterspeeches for $3,583$ hate speech instances. Further, we propose \model, a \textit{novel two-phased counterspeech generation framework}. In the first stage, \model\ learns vector-quantized representations for every intent and leverages the learned representations to generate desired intent-specific counterspeech in the second stage. Our comparative analysis and human evaluation demonstrate \model's superior performance over several baselines both empirically and qualitatively.

In brief, we make the following contributions:
\begin{itemize}[leftmargin=*,noitemsep,topsep=0pt]
    \item {\bf Novel task} -- {\em Intent-specific counterspeech generation}, which results in a diverse pool of counterarguments for a given hate speech.
    \item {\bf Novel dataset} -- \dataset with $6831$ counterarguments for $3583$ hate speeches spanning across five counterspeech intents.
    \item {\bf Novel model} -- \model, a two-phased intent-specific counterspeech generation framework.
    \item {\bf Evaluation} -- An extensive comparison and human evaluation to quantify the efficacy of our approach w.r.t state-of-the-art baselines.
\end{itemize}

\paragraph{Reproducibility:} We open-source the code and dataset at: \url{https://github.com/LCS2-IIITD/quarc-counterspeech}.


\section{Related Works}
\label{rel}

\paragraph{Counterspeech Datasets:}
An effective counterspeech can de-escalate the conversation and positively affect the audience of the counterspeech \cite{benesch2016considerations}. However, the scale limitations in manual counterspeech generation have prompted the automatic generation of counterspeech.
The first bottleneck is the availability of hate speech-counterspeech (\texttt{HS--CS}) datasets of high quality. Several strategies have been employed for dataset curation. \newcite{qian2019benchmark} focused on a crowd-sourcing approach in which non-expert crowd-workers were instructed to write responses to hate speeches from Reddit and Gab. The first large-scale \texttt{HS--CS} dataset, CONAN \cite{chung2019conan}, ensured quality by relying on niche-sourcing NGO experts to generate counterspeech. Further, to address the shortcomings of manual curation of datasets, \newcite{tekiroglu2020generating} presented a hybrid approach of dataset curation in which language models are trained on seed datasets of \texttt{HS--CS} pairs to generate new pairs validated and edited by annotators. Recently, \newcite{fanton-etal-2021-human} created Multi-Target CONAN, which contains labels for different target communities, and the counterspeeches are generated through a semi-automatic mechanism. 

\paragraph{Automatic Counterspeech Generation:} 
\newcite{qian2019benchmark} made an initial attempt to automatically generate counterspeeches using a Seq2Seq model. \newcite{zhu2021generate} employed a three-task pipeline consisting of an encoder, grammar check, and counterspeech retrieval based on hate speech for generating diverse counterspeeches. 
While research has shown the potency of using conditioned counterspeech depending on the context \cite{mathew2019thou, hangartner2021empathy}, the generation task is still in its infancy. Recently, \newcite{ijcai2022p716} proposed CounterGEDI, a model to control attributes like politeness, detoxification, and emotions of the generated counterspeeches using class-conditioned language models. However, the model does not include specific intents described in \newcite{benesch2016considerations}.

\paragraph{Controlling Methods for Generation:}
Prior studies on controlled language generation aimed to enforce user-specified constraints while generating texts. These approaches can exploit constraints at inference time \cite{DBLP:conf/iclr/DathathriMLHFMY20} or be applied during the training of the model \cite{wu2020controllable}. 
For controlled dialogue generation, \newcite{lin2021Adapter} used a series of lightweight adapters on top of a language model for high-level control of dialogues generated. In other work, \newcite{DBLP:journals/corr/abs-1909-05858} fine-tuned separate models for each attribute. While the above models show promising results for the task of controlled generation, we find that these models cannot be used directly for generating controlled counterspeeches with hate speech and intent as the input. This is due to the scarcity of counterspeeches for each intent and the overlap between the intents that make it harder for the model to learn the differences. 

\section{Dataset}
\label{data}
We begin by analyzing existing works to determine the intent categories for \dataset. CONAN \cite{chung2019conan} derives nine intent categories from \newcite{benesch2016considerations}, whereas \newcite{mathew2019thou} defined seven intent categories with minor variation. In contrast, due to the scarcity of data points, we club a few semantically-similar intents together in \dataset, e.g., we combine the \textit{positive} and \textit{affiliation} intents as \textit{positive}. In total, we consider five intent categories, i.e., \textit{informative}, \textit{question}, \textit{denouncing}, \textit{humor}, and \textit{positive} in this work. Table \ref{tab:intents} in Appendix \ref{app:annotation} highlights the relationship among the three sets of intent categories. 

The publicly-available Multi-Target CONAN dataset \cite{fanton-etal-2021-human} consists of \textasciitilde$5,000$ \texttt{HS--CS} pairs. However, it does not comprise any intent label for the counterspeeches. First, we extract the HS-CS pairs and clean them to remove redundancy. Subsequently, we employ three domain experts\footnote{The annotators are experts in NLP and social media.} to annotate the existing CS with an intent and write new CS for the remaining intent categories. Although, we capped the annotations at the values mentioned in Table \ref{data-target} so as to not induce repetitiveness; i.e. not every hate speech in IntentCONAN has five counterspeeches. The count-wise statistics are:  5 CS-per-HS: ~10\%,  4 CS-per-HS: ~5\%, 3 CS-per-HS: ~20\%, 2 CS-per-HS: ~10\%, and 1 CS-per-HS: ~55\%. An example of annotated counterspeeches for various intents is shown in Table \ref{tab1}.

\begin{table}[t!]
\centering
\resizebox{\columnwidth}{!}{
\begin{tabular}{l|c|c:c:c:c:c|c}
 \multicolumn{2}{c|}{\bf Hate Speech} & \multicolumn{6}{c}{\textbf{Counterspeech Intents}}\\\cline{1-8}
\textbf{Targets} & \textbf{Counts} & \textbf{INF} & \textbf{QUE} & \textbf{DEN} & \textbf{HUM} & \textbf{POS} & \textbf{Total} \\
\hline \hline
\textbf{Muslims} & 968 & 671 & 450 & 255 & 107 & 265 &1748\\
\textbf{Migrants} & 642 & 453 & 241 & 134 & 107 & 165 &1100 \\
\textbf{Women} & 517 & 415 & 225 & 195 & 158 & 158 &1151 \\
\textbf{LGBT+} & 465 & 280 & 195 & 145 & 99 & 132 &851 \\
\textbf{Jews} & 408 & 272 & 184 & 109 & 96 & 112 &773 \\
\textbf{POC} & 294 &  226 & 136 & 118 & 71 & 71 &622 \\
\textbf{Disabled} & 173 & 114 & 45 & 44 & 25 & 61 &289 \\
\textbf{Other} & 116 & 85 & 66 & 51 & 41 & 54 &297 \\
\hline
\textbf{Total} & 3583 & 2516 & 1542 & 1051 & 704 & 1018 & 6831 \\
\hline
\textbf{Train} & 2508 & 1761 &1079 &735 &494 &712 &4781 \\
\textbf{Dev} & 716 & 507 &310 &212 &139 &205 &1373 \\
\textbf{Test} & 359 & 248 &153 &104 &71 &101 &677 \\
\hline
\end{tabular}}
\caption{Statistics of \dataset. For each HS instance, \dataset\ has two CS on average.}
\label{data-target}
\end{table}

\begin{table}[t!]
\centering
\renewcommand{\arraystretch}{1.1}
\resizebox{\columnwidth}{!}{
\begin{tabular}{llp{8.5cm}}
\toprule
\textbf{HS} & \multicolumn{2}{l}{\multirow{2}{10cm}{The Jews did: USS Liberty Attack, the King David hotel attack, New York 9/11, London 7/7}} \\ \\
\midrule
\multirow{12}{*}{\textbf{CS}} & \multirow{2}{*}{\textbf{INF}} & \textit{Al-Qaeda had claimed responsibility for the London and New York attacks. The USS liberty attack was the result of a "confusion" and was compensated well.}\\ \cdashline{2-3}
& \textbf{QUE}$^{\dagger}$ & \textit{Where is the proof of this? Looks like assumptions as opposed to facts...}\\ \cdashline{2-3}
& \textbf{DEN} & \textit{This is just an unfounded conspiracy theory that only harms people.}\\ \cdashline{2-3}
& \multirow{2}{*}{\textbf{HUM}} & \textit{You must also believe that Earth is flat, moon landing is fake and we are actually ruled by lizard people and Illuminati, right?}\\ \cdashline{2-3}
& \multirow{2}{*}{\textbf{POS}} & \textit{Terrorism, as awful as it is, will not be solved if all we do is point fingers at the wrong person. United we stand, divided we fall.}\\
\bottomrule
\end{tabular}}
\vspace{-3mm}
\caption{Example of an annotated instance in \dataset. $^{\dagger}$Pre-existing counterspeech in the Multi-Target CONAN dataset \citep{fanton-etal-2021-human}.}
\label{tab1}
\end{table}

\paragraph{Annotation Guidelines:} 
Prior to the annotation, we make sure that the annotators have a comprehensive understanding of the field-manual\footnote{\url{https://onlineharassmentfieldmanual.pen.org/}.} for “responding to online abuse”. In our pilot study, we conduct several rounds of deliberation with all annotators over the understanding of the counterspeech. In particular, annotators consider the following objectives for every intent of speech:
    \textbf{Establish the Goal:} Each type of counterspeech necessitates a distinct fundamental idea, speech style, and goal.
    \textbf{De-escalate: } Each counterspeech instance should be written in a manner that would neutralize the situation and, ideally, not provoke retaliation or further hate speech.
    \textbf{Avoid Hostile Language: } Under no circumstance was threatening speech, name-calling, profanity, or hostility to be displayed while annotating counterspeech instances.
\begin{figure*}[ht!]
    \centering
    \includegraphics[width=\textwidth]{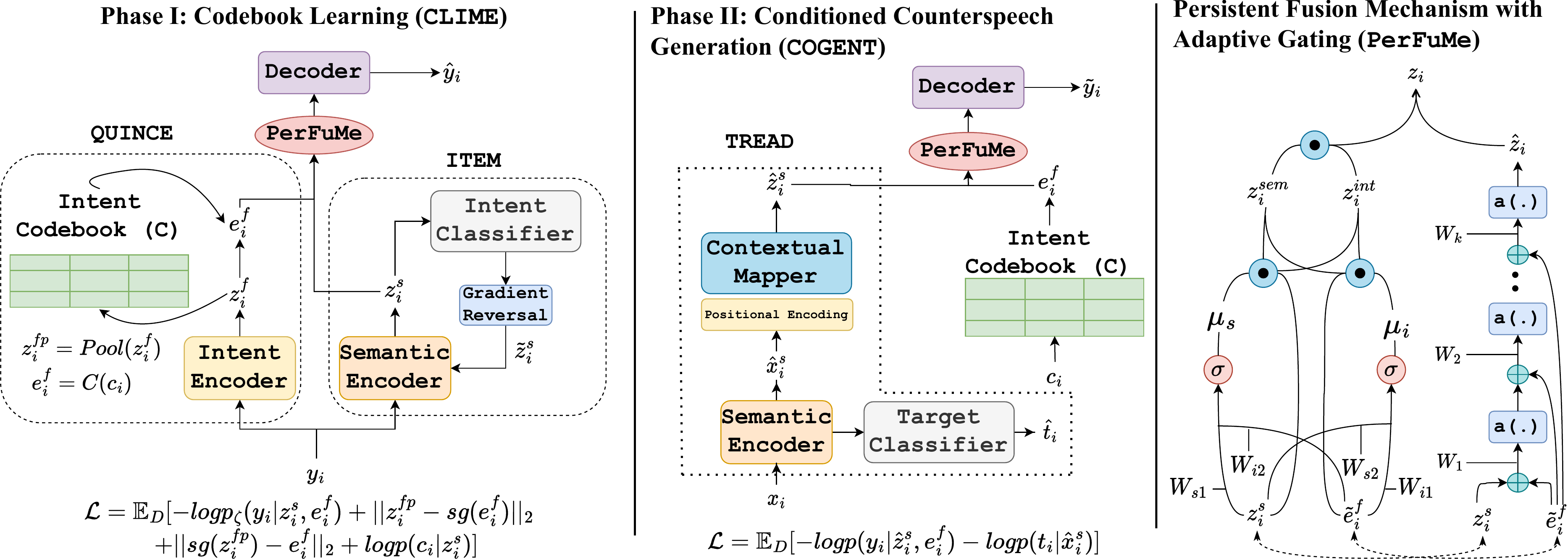}
    \caption{Architecture of our proposed framework, \model, which consists of two phases. The first-phase model, \Pa  is composed of two core modules, \Paa and \Pab, which are synchronised through the fusion module, \fusion to learn the Intent Codebook via reconstruction. The second-phase model, \Pb, uses \Pba to learn the contextual semantic mapping from hate speech to counterspeech, and fuses it with the intent vector from the learnt Intent Codebook using \fusion to generate intent-conditioned counterspeeches.}
    \label{fig:arch}
\end{figure*}
Subsequently, annotators label and write the intent-specific counterspeeches for $3,583$ distinct hate speech instances.  
Table \ref{data-target} shows \dataset 's detailed statistics. Appendix \ref{app:annotation} contains more information about the dataset.

\section{Proposed Methodology}
In this section, we define the architecture and the structural details of our proposed framework, \model. Our key insight is that a counterspeech instance can be decomposed into two distinct components -- its semantics and intent. In particular, we can convey the  semantics of the same counterspeech (which can be regarded as the compositional message) in multiple manners, such as through humor, as a question, in an informative manner, etc., depending upon the desired intent. More formally, given the counterspeech $y_i$, the semantics $s_i$ and the intent $c_i$, we posit that there exists a function $\zeta$ such that $y_i$ admits a factorization $y_i \sim \zeta(y_i|s_i, c_i)$. The primary goal of our method is to learn contextually-rich representations to seamlessly integrate the desired intent information with the semantics of the counterspeech to yield effective intent-conditioned counterspeeches. To this end, we design a novel two-phase training pipeline in which we attempt to learn the vector-quantized representations of each intent and propose a fusion mechanism, \fusion, to integrate this information into the model. 

Let us denote the dataset {$D = \{(x_1, t_1, c_1, y_1), \cdots, (x_n, t_n, c_n, y_n)\}$}, where $x_i$ denotes the $i^{th}$ hate-speech instance, $t_i$ denotes the target of $x_i$, $y_i$ denotes the counterspeech corresponding to $x_i$, and $c_i$ denotes the category/intent of $y_i$. Our end goal is to learn a stochastic counterspeech generation function $\chi$, such that $y_i \sim \chi(\cdot|x_i, c_i)$. We decompose this task into two phases, where we design two models: \Pa and \Pb. \Pa is designed to learn the quantized codebook vectors corresponding to each intent. This is done by learning a functional mapping $\zeta$, which aims to reconstruct the counterspeech $y_i$ from its semantic encoding $z_i^{s}$ and the intent encoding $e_i^{f}$ corresponding to $c_i$ as $\hat{y_i} \sim \zeta(\cdot |z_i^{s}, e_i^{f})$. For \Pb, we utilize the Intent Codebook $C$, assimilated through \Pa to learn $\chi$, which takes as input the semantic encoding of the hate speech $x_i^s$, as well as the encoding of desired intent $e_i^f$, to yield $\tilde{y_i} \sim \chi(\cdot|x_i^s, e_i^f)$. The overall architecture is depicted in Figure \ref{fig:arch}.

\subsection{Codebook Learning Model (\Pa)}
The overall purpose of \Pa is to learn the codebook representations for each intent category. It comprises  two modules: \Paa and \Pab; \Paa is utilized to generate the semantic encoding, while \Pab is utilized to procure the representation of the desired intent. The representations obtained from these modules are passed through our novel fusion mechanism, \fusion, and the emitted output is passed onto the decoder for the reconstruction of the original counterspeech. Note that \Pa does not utilize the hate speech instance $x_i$, and solely works on the counterspeech $y_i$ and its intent $c_i$ in a reconstructive fashion.

\paragraph{Intent-Unaware Semantic Encoding Module (\Paa):}
The counterspeech $y_i$ is first tokenized into its sub-word embeddings $\R{y_i^t}{n\times D}$, where $n$ is the maximum input length and $D$ is the latent dimension of the model. These embeddings are then passed through the semantic encoder, $\phi_s$, which is parameterized by a BART encoder, to yield the semantic representation $\R{z^i_s \sim \phi_s(\semy|y_i^t)}{n\times D}$. It is crucial that the information contained in $\semy$ reflects \textit{only the semantics} of the counterspeech, and not the intent, in order to enable effective learning of intent representations separately. If the intent information were distilled within $\semy$, the model would not need to rely on the codebook vector $\cvec$ to reconstruct the sample, rendering the learned intent distribution trivial. To combat this, we train an intent classification module on top of $\semy$, and use a gradient-reversal layer to expunge intent-specific information from within $\semy$. The intent classifier is trained jointly with the reconstruction module.

\paragraph{Quantized Intent Encoding Module (\Pab):}
The tokenized embedding $y_i^t$ is passed to the intent encoder, $\phi_i$ (parameterized by a BART encoder), to obtain the form encoding, $\syny \sim \phi_i(\syny|y_i^t)$. To learn a globally applicable quantized distribution for all intents, we employ a codebook similar to a VQ-VAE \cite{NIPS2017_7a98af17}. The intent-codebook, $\R{C}{|C| \times D}$, is a matrix where each row corresponds to the embedding of one intent. Our aim is to jointly learn the codebook for further utilization in generating intent-conditioned counterspeeches. We accomplish this by using the reconstruction objective as well as using a loss function similar to \newcite{NIPS2017_7a98af17}, which moves the pooled version of $\syny$ closer to the codebook vector $\cvec$ corresponding to $c_i$ ($\cvec = C(c_i)$), and vice versa, using a stop-gradient operator, $sg(.)$. $sg(.)$ is defined as identity and zero during forward and backward propagation, respectively. Since the semantic encoding $\semy$ has had its intent-specific information stripped through the gradient reversal layer, this information must be distilled in the quantized $\cvec$ in order to facilitate effective reconstruction.
\vspace{-0.4em}
\paragraph{Reconstruction:}
The generated embeddings $\semy$ and $\cvec$ (from \Paa and \Pab, respectively) are then passed into our adaptive-gated fusion mechanism, \fusion to yield $\R{z_i}{n \times D}$. $z_i$ is then given to the decoder as input to generate $\hat{y_i} \sim \zeta(\cdot|\semy, \cvec)$, the reconstructed output. We train the model by minimizing the negative log-likelihood of $\hat{y_i}$ with respect to the reference $y_i$ as well as incorporating auxiliary losses from \Paa and \Pab, as follows:
\begin{multline}
  \label{eq:l1}
\mathcal{L} = \mathbb{E}_{D}[-logp_{\zeta}(y_i|\semy, \cvec) + ||z_i^{fp}-sg(e_i^f)||_2 + \\
||sg(z_i^{fp})-e_i^f||_2 + logp(c_i|\semy)]
\end{multline}
where $\semy \sim \phi_s(\cdot|y_i)$, $\cvec = C(c_i)$, $\syny \sim \phi_i(\cdot| y_i)$ and $z_i^{fp}=Pool(\syny)$.

\subsection{Conditioned Counterspeech Generation Model (\Pb)}

The objective of the second phase is to generate counterspeeches that are conditioned on the desired intent, given an input hate speech. This is achieved through the utilization of \Pb, which comprises  \Pba, a module designed to map the input hate speech $x_i$ to a semantic encoding of the counterspeech, which can then be fused with the codebook vector $\cvec$ corresponding to the specified intent as learned through \Pa. The following sections provide a more in-depth description of the functions of these modules.

\paragraph{Target-Aware Semantic Mapping Module (\Pba):}
The hate speech $x_i$ is passed through the semantic encoder $\phi_s$ to obtain its semantic representation $\R{\semx \sim \phi_s(\cdot|x_i)}{n \times D}$. \\
\textit{-- Target Information Incorporation:} Since the semantics of the hate speech should inherently possess discriminative characteristics to determine the intended target of hate speech, we explicitly strengthen $\semx$ by incorporating target category $t_i$ through a joint classification loss. $\semx$ is passed through a target classification module to yield $\R{\trgx}{|T|}$, where $|T|$ denotes the total number of target categories in the dataset. $\trgx$ denotes the probability distribution over all targets for $x_i$ and is optimized via the negative log-likelihood loss with the actual target $t_i$. \\
\textit{-- Semantic Mapping:} The semantic representation $\semx$ encompasses information about the semantics of hate speech; however, we require the semantics of the corresponding counterspeech to coalesce with the desired intent. To facilitate this, we define a mapping function $\xi$, which maps the semantics of hate speech to the desired counterspeech as $\mapx \sim \xi(\cdot|\semx)$. In practice, $\xi$ is parameterized by a multi-layered Transformer Encoder \cite{NIPS2017_3f5ee243}, which is learned jointly. We term the parameterized version of $\xi$ as the contextual mapper.

\paragraph{Counterspeech Generation:}
The semantic mapping of  counterspeech, $\R{\mapx}{n \times D}$ is then fused with the codebook vector $\cvec$ through \fusion and passed to the decoder to yield the generated counterspeech $\R{\tilde{y}_i \sim \chi(\cdot|\mapx, \cvec)}{n \times D}$. \Pb is trained by minimizing the negative log-likelihood loss of generating $y_i$, as well as the auxiliary target loss as follows:
\begin{equation}
  \label{eq:l2}
  \begin{gathered}[b]
\mathcal{L} = \mathbb{E}_{D}[-logp(y_i|\mapx, \cvec) - logp(t_i|\semx)]
  \end{gathered}
\end{equation}
with $\semx \sim \phi_s(\cdot|x_i), \mapx \sim \xi(\cdot|\semx)$ and $\cvec = C(c_i)$.

\subsection{Persistent Fusion Mechanism with Adaptive Gating (\fusion)}
Coalescing intent-specific information  with the semantics of a counterspeech can prove to be a challenging task as the model may not pay heed to the desired intent and generate a counterspeech that respects the desired semantics but has a different form than required. To address this problem, we propose \fusion, a persistent fusion module where we repeatedly synchronize the intent-encoded information with the semantic information to ensure that the desired form is not overlooked. We also enhance this fusion procedure with adaptive gating, where we design two distinct gates to control the degree of semantic and intent-specific information leveraged during integration.

More formally, let the semantic and intent-specific information be denoted by $\R{\semy}{N \times D}$ and $\R{\cvec}{1 \times D}$, respectively. $\cvec$ is stacked on top of itself $N$ times to obtain $\R{\ivec}{N \times D}$. We obtain $\R{\hat{z}_i}{N \times D}$ as:
\begin{multline}
  \label{eq:f1}
    \hat{z}_i = a( \dots a((a((\semy \oplus \ivec)W_1 + b_1) \oplus \ivec)W_2 + b_2) \\ \dots \oplus \ivec)W_k + b_k)
\end{multline}
where $a$ denotes a non-linear activation function, $\oplus$ represents concatenation, $\R{W_1, W_2 \dots W_k}{2D \times D}$, and $\R{b_1, b_2 \dots b_k}{N \times D}$ are trainable matrices. We also introduce two gates, \textit{s-gate} and \textit{i-gate}, which control the flow of semantic and intent-specific information, respectively.
\begin{eqnarray}
    \mu_s & = & \sigma(\semy W_{s1} + \ivec W_{i2} + b_{s}) \nonumber \\
   \mu_i & = & \sigma(\semy W_{s2} + \ivec W_{i1} + b_{i}) 
\end{eqnarray}
$\R{W_{s1}, W_{s2}, W_{i1}, W_{i2}}{D \times D}$, and $\R{b_s, b_i}{N \times D}$ are trainable parameters. $\mu_s$ and $\mu_i$ are designed to filter the information emitted from the semantic and intent-specific encodings, respectively. 
\begin{eqnarray}
    z_i^{sem} & = & \mu_s \odot \semy + (1-\mu_s) \odot \ivec \nonumber \\
    z_i^{int} & = & (1-\mu_i) \odot \semy + \mu_i \odot \ivec
\end{eqnarray}
where $\odot$ denotes the Hadamard product. Finally, we resolve the information obtained from \textit{s-gate} ($z_i^{sem}$), \textit{i-gate} ($z_i^{int}$) and the persistent fusion mechanism ($\hat{z}_i$) to produce the fused matrix $z_i = \hat{z}_i + z_i^{sem} \odot z_i^{int}$, where $\R{z_i}{N \times D}$.

\section{Experimental Setup and Results}\label{exp} 
In this section, we delineate an  exhaustive analysis of our model's performance and also carry out a predictive comparison against text generation models using both human and automatic evaluation. 

\paragraph{Comparative Systems:}\label{base}
$\bullet$ \textbf{Generate Prune Select (GPS) }\cite{zhu2021generate} uses a three-stage pipeline for generating counterspeeches. The first stage generates a large number of counterspeeches using an autoencoder architecture which is further pruned using a grammatical model. Finally, the most suitable counterspeeches are chosen for hate speech using a vector-based response selection model.
$\bullet$ \textbf{Plug And Play Language Model (PPLM)} \cite{DBLP:conf/iclr/DathathriMLHFMY20} 
We utilize fine-tuned GPT-2 as the base language model for PPLM. 
$\bullet$ In addition, we fine-tune \textbf{DialoGPT} \cite{zhang-etal-2020-dialogpt} and \textbf{BART} \cite{lewis2019bart} on \dataset\ as well. For all four comparative models, we provide the desired intent as prompt. 

\paragraph{Evaluation Metrics:}\label{results}
We employ \textit{Rouge} \cite{lin-hovy-2003-automatic} and \textit{Meteor} \cite{banerjee2005meteor} scores to evaluate the syntactic correctness of the generated counterspeech. Given that Rouge and Meteor metrics primarily assess surface-level overlap, their standalone usage may not provide a comprehensive evaluation of the effectiveness of the generated counterspeech instances, considering the possibility of multiple correct outputs. To address this limitation, we augment these metrics by incorporating measures of semantic richness and conducting thorough human evaluations to ensure a more comprehensive assessment. For semantic richness, we report \textit{BERTScore} (BS) \cite{DBLP:conf/iclr/ZhangKWWA20} along with  \textit{cosine similarity} (SS) obtained from a sentence-transformers model (all-miniLM-v2) \cite{reimers-gurevych-2019-sentence}. Moreover, to check the efficacy of the models in incorporating the desired intent in the generated counterspeeches, we compute \textit{category accuracy} (CA) through an intent classification ($IC$) model.

\begin{table}[t!]
\centering
\resizebox{\columnwidth}{!}{
\begin{tabular}{l|c|c|c|c|c|c|c}
\toprule
\multirow{2}{*}{\textbf{Method}} &  \multicolumn{3}{c|}{\textbf{ROUGE}} & \multirow{2}{*}{\textbf{M}} & \multirow{2}{*}{\textbf{SS}} &  \multirow{2}{*}{\textbf{BS}} &   \multirow{2}{*}{\textbf{CA}} \\ \cline{2-4}
 & \textbf{R1} &  \textbf{R2} &  \textbf{RL} &  & & &  \\ \hline
DialoGPT &  \multicolumn{1}{r|}{0.13} &  \multicolumn{1}{r|}{0.01} &  0.11 &  0.15 &  0.65 &  0.81 &  0.34  \\ 
BART &  \multicolumn{1}{r|}{0.17} &  \multicolumn{1}{r|}{0.04} &  0.16 &  0.16 &  0.72 &  0.87 &  0.65  \\ 
PPLM &  \multicolumn{1}{r|}{0.15} &  \multicolumn{1}{r|}{0.02} &  0.13 &  0.14 &  0.72 &  0.82 &  0.33  \\ 
GPS &  \multicolumn{1}{r|}{0.23} &  \multicolumn{1}{r|}{\textbf{0.10}} &  0.21 &  0.19 &  0.73 &  0.87 &  0.39 \\ 
\hline
\textbf{\model} &  \multicolumn{1}{r|}{\textbf{0.25}} &  \multicolumn{1}{r|}{0.08} &  \textbf{0.24} &  \textbf{0.22} &  \textbf{0.77} &  \textbf{0.89} &  \textbf{0.70} \\ \hline
- CI &
  \multicolumn{1}{r|}{0.23} &  \multicolumn{1}{r|}{0.06} &  0.22 &  0.21 &  0.77 &  0.88 &  0.66 \\ 
- \Pa &  \multicolumn{1}{r|}{0.22} &  \multicolumn{1}{r|}{0.06} &  0.19 &  0.20 &  0.73 &  0.86 &  0.69  \\ 
- \fusion &  \multicolumn{1}{r|}{0.18} &  \multicolumn{1}{r|}{0.04} &  0.17 &  0.16 &  0.68 &  0.83 &  0.64 \\ 
- Residual &  \multicolumn{1}{r|}{0.18} &  \multicolumn{1}{r|}{0.04} &  0.15 &  0.16 &  0.68 &  0.84 &  \textbf{0.70} \\ 
+ MB &  \multicolumn{1}{r|}{0.16} &  \multicolumn{1}{r|}{0.03} &  0.13 &  0.14 &  0.67 &  0.84 &  0.68 \\ 
$k=1$ &  \multicolumn{1}{r|}{0.25} &  \multicolumn{1}{r|}{0.08} &  0.24 &  0.22 &  0.76 &  0.89 &  0.66 \\ 
$k=5$ &  \multicolumn{1}{r|}{0.25} &  \multicolumn{1}{r|}{0.08} &  0.24 &  0.21 &  0.77 &  0.89 &  \textbf{0.70} \\ 
\bottomrule
 \end{tabular}}
\caption{Comparative results for \model. CI: Codebook Initialization; MB: Memory Bank.}
\label{tab:results}
\end{table}

\paragraph{Result Analysis:}
The results are reported in Table \ref{tab:results}. We observe that \model beats the baselines across all metrics. In terms of lexical similarity, GPS is the best-performing baseline as it demonstrates high scores on R1, R2, RL, and Meteor. However, \model reports higher scores by a margin of \textasciitilde10\% on the syntactic similarity measures except for R2. On the semantic similarity measure, \model outperforms the best baseline (GPS) by \textasciitilde2\% and \textasciitilde5\% on BS and SS scores, respectively. This demonstrates the ability of our framework to generate semantically coherent counterspeeches to a given hate speech. In the context of generating intent-conditioned counterspeeches, CA evaluates the appropriateness of the generated counterspeeches. We observe that the majority of the baselines are notably inferior in producing outputs corresponding to the desired intent. For instance, while GPS is able to produce syntactically and semantically coherent outputs, it falls short in terms of accurately preserving the intended intent and is outperformed by our framework by 79\%. Due to the explicit design of our pipeline, \model is able to efficaciously generate counterspeeches that preserve the desired intent (c.f. Appendix \ref{appendix:analysisIC}). 

\begin{table}[h]
\begin{center}
\begin{tabular}{c|c|c} 
 \hline
\textbf{Method} & \textbf{Div} & \textbf{Nov} \\ \hline
GPS                   & 0.36          & 0.33          \\ \hline
BART                  & 0.42          & 0.62          \\ \hline
\model & \textbf{0.68} & \textbf{0.67} \\ \hline
\end{tabular}
\caption{Analyzing lexical dissimilarity w.r.t. novelty and diversity scores.
}
\label{tab:nov-div}

\end{center}
\end{table}

To obtain a deeper insight into the performance of \model and the best baselines (GPS and BART), we compute \textit{novelty} and \textit{diversity} in line with \newcite{ijcai2018p618} (c.f. Table \ref{tab:nov-div}). These metrics measure the lexical dissimilarity between the generated instances and the training corpus, as well as the dissimilarity between the generated instances themselves. They convey the degree of originality and variety in the generated text and can serve as an indicator of over-reliance on memorization of the training corpus or repetition of outputs. The scores achieved by \model (0.68 and 0.67) indicate a relatively higher degree of originality and variety in the generated text, while GPS exhibits the lowest scores across these metrics, indicating a tendency to memorize data from the training set and generate similar outputs more frequently. This might also be one of the reasons for its relatively high performance on similarity metrics. We must observe, however, that a high score on these metrics, when accompanied by a low semantic and syntactic similarity, marks that the generated outputs may not be entirely aligned with the intended ones.


\begin{table*}[ht!]
\centering
\renewcommand{\arraystretch}{1.1}
\resizebox{\textwidth}{!}{
\begin{tabular}{c|p{3.6cm}|c|p{13cm}|c|c}
\hline
& \multicolumn{2}{c|}{\bf Inputs} & \multicolumn{1}{c|}{\bf Output} & \multicolumn{2}{c}{\bf Evaluation} \\ \cline{2-6}
& \multicolumn{1}{c|}{\bf Hate Speech} & \bf Desired CS Intent & \multicolumn{1}{c|}{\bf Counterspeech} & Intent & CS \\
\hline
\multirow{7}{*}{H1} & \multirow{7}{4cm}{Every Muslim is a potential terrorist.} & \multirow{3}{*}{Positive} & \textbf{\model:} \em Islam is a religion of peace and love. Muslims are not terrorists. They are just like us. & Positive & Yes\\ \cdashline{4-6} 
& & & \textbf{Best Baseline:} \em How can you say a person is a potential terrorist? & \textcolor{red}{Question} & Yes\\ \cline{3-6}
 &  & \multirow{4}{*}{Denouncing} & \em \textbf{\model :}What you are saying has nothing to do with Muslims. You are wrong to think that all Muslims are terrorists. & Denouncing & Yes\\ \cdashline{4-6} 
& & & \textbf{Best Baseline:} \em not all muslims are muslims. ignoring all muslims are peaceful citizens, but some people are terrorists. & Denouncing & \textcolor{red}{No}\\
\hline
\multirow{7}{*}{H2} & \multirow{7}{4cm}{Women are basically childlike, incapable of serious thought and emotion.} & \multirow{4}{*}{Informative} & \textbf{\model :} \em There is no evidence that women are "childlike" or "ignorant". Women are human beings with feelings and emotions. & Informative & Yes\\ \cdashline{4-6} 
& & & \textbf{GPS:} \em the problem is founded on the age of women to article rights, and society. why is it possible for those us world take a world?. & \textcolor{red}{Question} & \textcolor{red}{No} \\ \cline{3-6}
& & \multirow{3}{*}{Humor} & \textbf{\model : } \em So you think women are "childlike" because they are incapable of thinking? & \textcolor{red}{Question} & Yes \\ \cdashline{4-6}
& & & \textbf{GPS:} \em Of course a woman won't remain a child most of her life. Otherwise you can't call them women. Your statement just reflects the patriarchy & \textcolor{red}{Denouncing} & Yes \\ 
\bottomrule
\end{tabular}}
\caption{Qualitative evaluation. The intent and CS columns are appropriately labelled by human experts to assess the validity of outputs corresponding to the input intent and relevance to the specific hate speech.}
\label{tab:tab-error}
\end{table*}
\vspace{-0.4em}
\paragraph{Ablation Study:}
To dissect and identify the individual components that drive our framework, we perform multiple ablations on its architecture. In the standard version of \model, we initialize the codebook vectors by using the mean-pooled version of the representations of each intent obtained from an intent classifier. When we remove this initialization strategy and initialize the codebook vectors randomly for \Pa, we observe a slight drop-off in all metrics. However, when we retain this initialization strategy and directly use these vectors in \Pb without undergoing the first phase, a higher drop in almost all scores (except CA) is observed. The drop is especially high in diversity, which demonstrates that the generated texts have more repeated tokens across test samples, explaining the higher CA score as compared to the first ablation. 

We performed another ablation in which we added a memory bank component to \Pb. We stored the semantic representations $\semy$ of each counterspeech instance in training set in a memory bank in the first phase while utilizing \Pa. When we perform contextual mapping in the \Pba module inside \Pb to map the semantics of the hate speech $\semx$ to that of the corresponding counterspeech $\mapx$, we used the representations stored in the memory bank to align $\mapx$ and $\semy$ closer to each other via an auxiliary loss given by $||\semy - \mapx||_2$. However, this ended up degrading the performance, perhaps due to the overfitting and lack of generalization owing to the relatively smaller training set size. We performed another ablation in which we removed all residual connections from both \Pa and \Pb to see its effect, and we noted a similar drop in performance, In the last two ablations, we again noted a large drop in diversity, which demonstrates that both \Pa and residual connections are critical in generating non-repetitive distinct counterspeeches.

\paragraph{Qualitative Analysis:}
For qualitative evaluation, we report the outputs of \model\ and the best baseline (GPS) for two instances in Table \ref{tab:tab-error}. In each case, we show the outputs for two desired CS intents. We observe that \model\ does a fair job in generating CS with the desired intents in three out of four cases, whereas the intents of generated CS in GPS align with the desired intent in only one out of four cases -- even for the correct case, GPS produces an incoherent statement. For H2 with the desired \textit{humor} intent, both \model\ and GPS commit mistakes for the intent (i.e., \textit{question} for \model\ and \textit{denouncing} for GPS); however, the output is a valid CS, ignoring the desired intent. Our analysis suggests that GPS and other baselines perform poorly in generating the desired intent-conditioned CS as compared to \model.

\vspace{-0.3em}
\paragraph{Human Evaluation:}
Given the limitations of empirical evaluation in holistically assessing the efficacy of generation models, we conduct a comprehensive human evaluation on a random subset of the generated counterspeeches from \model\ and GPS (detailed instructions in Appendix \ref{appendix:humanEval}). The subset was uniformly distributed across intents. We ask our evaluators\footnote{A total of 60 evaluators in the field of NLP and social science participated, having ages between 20-30 years with 60:40 male to female ratio.} to rate the outputs on the following metrics: \textbf{Independent CS (IC)} denotes whether the generated instance can be considered as CS without any context; \textbf{Conditioned CS (CC)} shows whether the generated output is an appropriate response to the given hate speech; \textbf{Adequacy (A)} depicts whether the generated CS is grammatically sound, coherent and fluent; \textbf{Toxicity (T)} indicates whether the output can be considered toxic. For each of the above metrics, the evaluators are instructed to rate every counterspeech on a 5-point Likert scale. For example, considering the Toxicity metric (T),  a score of 1 denotes that the counterspeech can be considered completely non-toxic, 3 denotes neutral and 5 denotes highly toxic. \textbf{Category Accuracy (CA)} determines if the counterspeech adheres to the desired intent; here the evaluators are told to assign the counterspeech to one of the five intents to the best of their ability. 

\begin{table}[t]
\centering
\resizebox{0.9\columnwidth}{!}{
\begin{tabular}{c|c|c|c|c|c}
\hline 
\multirow{2}{*}{\bf Model} & \multicolumn{5}{c}{\textbf{Human Evaluation Metric}} \\
\cline{2-6}
& \textbf{IC $\uparrow$} & \textbf{CC $\uparrow$} & \textbf{A $\uparrow$} & \textbf{T $\downarrow$}  & \textbf{CA $\uparrow$} \\ \hline
\textbf{\model} & \textbf{3.69} & \textbf{3.76}  & \textbf{4.10} & 2.42 & \textbf{0.7} \\ \hline
\textbf{GPS} & 3.16  & 3.04 & 3.32 & \textbf{2.30} & 0.1 \\ \hline
\end{tabular}}
\caption{Human evaluation on 5-point Likert scale (except for CA, which represents the proportion of counterspeeches with matching intents as annotated by evaluators).}
\label{tab:human}
\end{table}


The results of the human evaluation (c.f. Table \ref{tab:human}) indicate that \model outperforms the best baseline by a significant margin in all metrics except toxicity. These results demonstrate that the outputs generated by our model are not only more effectively recognized as counterspeeches but are also more closely aligned with the intended response to the consumed hate speech. Moreover, the results attest to the efficacy of our intent-specific representation and fusion-based approach through the CA metric. We observe fair agreement ($\kappa=0.32$) on Fleiss' Kappa scale amongst the evaluators \cite{joseph1973Equivalence}.

\begin{figure}[t]
     \centering
     \begin{subfigure}[b]{0.48\columnwidth}
         \centering
         \includegraphics[width=\linewidth]{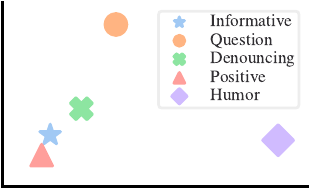}
         \label{fig:codebook-embed}
     \end{subfigure}
     \begin{subfigure}[b]{0.48\columnwidth}
         \centering
         \includegraphics[width=\linewidth]{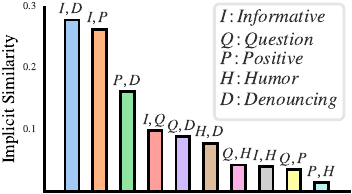}
         \label{fig:imp-sim}
     \end{subfigure}
     \vspace{-1.2em}
     \caption{\textit{Left:} A scatter plot of the codebook vectors (after dimensionality reduction) corresponding to different intents. \textit{Right:} The Implicit Similarity ($IS$) between intent pairs captured through human evaluation.}
     \label{fig:interpretability}
\end{figure}

\paragraph{Congruence:} We introduce Implicit Similarity $(IS)$, a metric that utilizes implicit feedback from human evaluation to reflect the similarity between intent pairs. Intuitively, the core idea behind $IS$ is that when different evaluators assign a different intent category to the same counterspeech, there exists a certain affinity between those categories. As an example, if evaluator $A$ assigns the intent $Informative$ to a counterspeech, and evaluator $B$ assigns the intent $Positive$ to the same counterspeech, then there exists a certain similarity between the intents $Informative$ and $Positive$. The strength of this affinity can be approximated via its relative frequency of occurrence, and the method for its computation is described below.

We calculate $IS$ for every possible intent pair; since there are $5$ intents, there are a total of $\Comb{5}{2}=10$ distinct pairs. Let the counterspeech $y_i$ be generated in response to the hate speech $x_i$ with the desired intent $c_i$. The human evaluators are asked to classify the intent of $y_i$ from the defined set of $5$ intents -- $\{I_1, I_2, I_3, I_4, I_5\}$ without knowledge of the actual intent $c_i$. Each evaluator from the group of $N$ evaluators assigns the intent for $y_i$ and we obtain the relative frequency of the classified intents as $V_i = \{I_1: v_{i1}, I_2: v_{i2}, I_3: v_{i3}, I_4: v_{i4}, I_5: v_{i5}\}$, where $\sum_{j=1}^5 v_{ij} = 1$, and $v_{ij}$ denotes the fraction of evaluators that assigned $y_i$ to the intent class $I_j$. The implicit similarity for a pair of intents $(I_a, I_b)$ for the $i^{th}$ counterspeech is computed as $IS_i^{a,b} = v_{ia}\times v_{ib}\times NS$, where $NS = 4$ is the normalizing factor applied to standardize the range of $IS_i^{a,b}$ to $[0,1]$ (since the maximum value of $v_{ia}\times v_{ib}$ is $0.25$). $IS_i^{a,b}$ is indicative of the similarity between a pair of intents, as a higher value of $IS_i^{a,b}$ deems that the same sample was assigned to both $I_a$ and $I_b$ consistently by evaluators (without knowledge of the desired $c_i$), and thus, there exists a certain affinity between these intent classes. Hence, we compute the overall implicit similarity between $(I_a, I_b)$ for the set of $K$ counterspeeches given to the human evaluators as $IS^{a,b} = \sum_{k=1}^K IS_k^{a,b}/K$. Note that $IS$ is calculated without the knowledge of the desired intent $c_i$ to provide a more faithful picture.

We plot the learnt representation of each intent category (after dimensionality reduction through PCA) along with the computed $IS$ scores (Figure \ref{fig:interpretability}). We note that the $IS$ scores \textit{closely align} with the distances between the learnt representations. This congruence not only demonstrates the robustness of the learnt representations, but also provides a key insight into a critical factor behind the superior performance of \model (more details in Appendix \ref{appendix:is}).



\section{Conclusion}\label{conclusion}
In an effort to address the pervasive issue of hateful speech on the internet, we proposed the novel task of intent-conditioned counterspeech generation. We developed \dataset, the first intent-specific dataset for diverse counterspeech generation. Further, to benchmark the dataset, we proposed a novel framework (\model) that decomposes the task into two phases -- \Pa\ learns the intent distribution which is subsequently leveraged by \Pb\ to generate the intent-conditioned counterspeeches. We conducted an extensive evaluation (i.e., empirical, qualitative, and human) to establish the effectiveness of \model. 
\section*{Acknowledgement}
Authors acknowledge the partial support of Logically and Infosys Center of AI (CAI), IIIT Delhi. 

\section*{Limitations}
The current work marks the first step towards intent-conditioned counterspeech generation, and as we noted, even though our model excels in fluency, a larger and more diverse dataset paired with knowledge grounding is necessary to improve and ensure factual correctness. Although the annotators kept the quality of counterspeech as high as possible, it is possible that this data is not at par with other datasets that are annotated by more skilled NGO operators, as is the case with the Multi-Target CONAN dataset \cite{fanton-etal-2021-human}. A more large-scale annotation of our dataset with higher instances for under-represented target communities would hence be beneficial to learn more accurate distributions of every counterspeech class. Another limitation of the current work is that it exhibits a slightly higher-degree of toxicity compared to the baseline. It, therefore, pertains to accounting for lowering the amount of toxicity present in the generated counterspeeches as future research. Lastly, humor in  counterspeech is a very subjective topic, and inspite of including only a few datapoints from that class as compared to the others in our dataset, it is likely that \model could generate vague and/or offensive text under the pretext of humor. We intend on keeping the dataset private and only provide access for research and educational purposes.  

\section*{Ethics Statement}\label{ethics}
We recognize that combating online hate speech can be a delicate matter, and we fully acknowledge that research in this domain might raise ethical and moral concerns. This work is simply the beginning of efforts to create a consistent and diversified compendium of counterspeeches  for every hateful instance. We also agree that models used to automate counterspeech could end up producing factually erroneous statements, and a more efficient method of incorporating real-world knowledge into these models is required. On the other hand, even if generative models could perform well, there is still a pressing need for a large-scale counterspeech dataset with a more diversified response pool to ensure a net positive outcome. Furthermore, while a deployable model for counterspeech is not completely feasible as of now, there are organizations like United Against Hate\footnote{\url{https://www.united-against-hate.org/}.} who are making considerable contributions to mitigate hate online.

\bibliography{anthology,custom}
\bibliographystyle{acl_natbib}

\newpage

\clearpage
\appendix
\section{Dataset}
\label{app:annotation}

\subsection{Objectives for the Annotators}
Following are the guidelines strictly adhered to by the annotators while constructing counterspeeches for the five intents:

\paragraph{Informative Counterspeech:}
Informative or Factual statements are the most popular form of counterspeech in which  a counterspeech seeks to debunk or fact-check the claims made by the attacker. We found some instances of counterspeech in the CONAN dataset which were inconsistent with the definition of facts, a few being:

\settowidth\widest{\textbf{word}}
    \begin{description}[leftmargin=\dimexpr\widest+\labelsep\relax,labelindent=0pt,
        labelwidth=\widest]
    \small
    \item[{\fontfamily{qcr}\selectfont({\small HS})}]
    \textbf{“Don't the Moslem's Holy books say 'kill infidels wherever they may be hiding'”}
    \item[{\fontfamily{qcr}\selectfont({\small CS})}]
    \textit{No, it doesn’t.} 
    \item[{\fontfamily{qcr}\selectfont({\small HS})}]
    \textbf{“Muslims are the ones who invented the slave trade.”} 
    \item[{\fontfamily{qcr}\selectfont({\small CS})}]
    \textit{Slavery has been rampant in early societies of all kinds. I am sure the Romans did not take inspiration for their slave trade directly from Muslims.}
    \end{description}

Therefore, to ensure the validity of the counterspeeches without having to fact-check every statement from the Multi-Target CONAN dataset, we decide to rename the class to "Informative Counterspeech"; this seems more apt, and makes writing counterspeeches for our annotators easier. Furthermore, because our annotators were unfamiliar with facts from each target community, we relied on official sources like Red Cross, The Holocaust Encyclopedia, \href{https://www.rainn.org/}{RAINN}, \href{https://www.adl.org/}{The Anti-Defamation League}, \href{https://www.brookings.edu/}{Brookings}, and credible news sources like  \href{https://edition.cnn.com/}{CNN},  \href{https://www.huffingtonpost.co.uk/}{HuffingtonPost} (among others) to verify that the annotations were factually correct as far as possible for this class.

\paragraph{Questioning Counterspeech:} For this class, the annotators were instructed to frame countermeasures in the form of questions that would challenge the speaker's chain of reasoning and compel them to either answer convincingly or recant their original remark. If necessary, factual information was to be obtained from a pre-determined pool of data sources, as indicated in the preceding section.

\paragraph{Denouncing Counterspeech:}
This category of counterspeech needed to be handled with caution, as denouncing can sometimes be used to propagate obscene language. Our annotators were directed to convey the impression that the opinions put forth by the hate speaker are not acceptable without using name-calling or profanity.

\paragraph{Humorous Counterspeech:}
A heated dispute or discussion can be effectively defused by humor and sarcasm \cite{mathew2019thou}. By highlighting how absurd it is, humor undercuts the hate speech and aids in diverting the attention of those following the dialogue online. Annotators were asked to construct a sentence that would not incite resentment from other users while also making sure that it would not contain any controversial ideas or terms. It should be mentioned that the annotators had prior knowledge of the sarcasm and humour that are well-received on social media.

\paragraph{Positive Counterspeech:}
The use of empathy and positive reinforcement in hate speech can lead to a decline in online animosity \cite{hangartner2021empathy}. Regardless of the severity of the hate speech, the annotators make an effort to compose a courteous, polite, and civil statement. Furthermore, we argue that if bystanders who are following the discourse online are a member of the group impacted by the comment, they would be instilled with a sense of support and humanness.

\begin{table}[t!]
\centering
\resizebox{0.48\textwidth}{!}{
\begin{tabular}{l|l|l}
\toprule
\newcite{benesch2016considerations} & \newcite{mathew2019thou} & \dataset \\ 
\midrule
Facts & Facts & Informative \\ \hline
Humor & Humor & Humor \\ \hline
Question & -- & Question \\ \hline 
Denouncing & Denouncing & \multirow{3}{*}{Denouncing} \\ \cline{1-2}
Consequences & Consequences & \\ \cline{1-2}
Hypocrisy & Contradictions & \\ \hline
Affiliation & Affiliation & \multirow{2}{*}{Positive} \\ \cline{1-2}
Positive & Positive & \\ \hline
Other & -- & -- \\ \hline
\end{tabular}}
\caption{Comparison of intent categories from existing works and \dataset.}
\label{tab:intents}
\end{table}

\subsection{Dataset Statistics}
\label{app:data}
\begin{figure}[h]
     \centering
     \begin{subfigure}[b]{0.45\columnwidth}
         \centering
         \includegraphics[width=\linewidth]{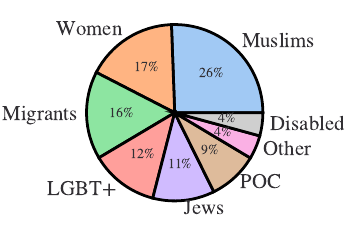}
         \caption{HS Target distribution}
         \label{fig:datastats1}
     \end{subfigure}\hspace{1em}
     \begin{subfigure}[b]{0.45\columnwidth}
         \centering
         \includegraphics[width=\linewidth]{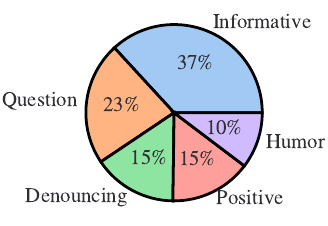}
         \caption{CS Intent distribution}
         \label{fig:datastats2}
     \end{subfigure}
     \\
     \begin{subfigure}[b]{0.45\columnwidth}
         \centering
         \includegraphics[width=\linewidth]{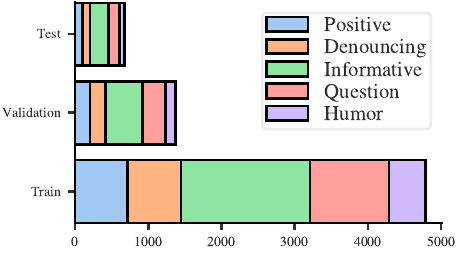}
         \caption{CS Intent distribution across datasets}
         \label{fig:datastats3}
     \end{subfigure}\hspace{1em}
          \begin{subfigure}[b]{0.45\columnwidth}
         \centering
         \includegraphics[width=\linewidth]{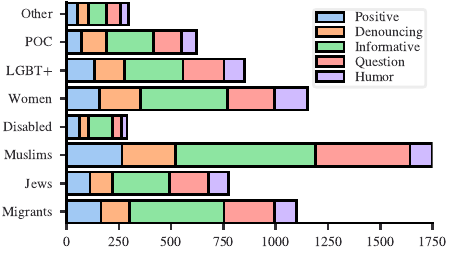}
         \caption{Intent and Target distribution}
         \label{fig:datastats4}
     \end{subfigure}
               \begin{subfigure}[b]{0.45\columnwidth}
         \centering
         \includegraphics[width=\linewidth]{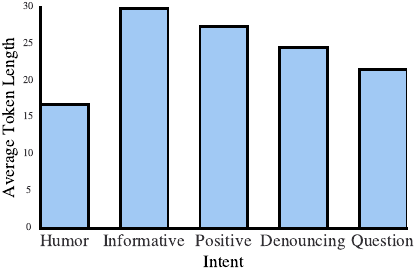}
         \caption{Intents: Mean token length}
         \label{fig:datastats5}
     \end{subfigure}
                    \begin{subfigure}[b]{0.45\columnwidth}
         \centering
         \includegraphics[width=\linewidth]{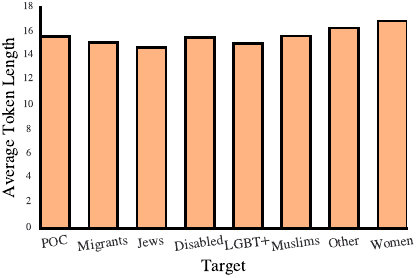}
         \caption{Targets: Mean token length}
         \label{fig:datastats6}
     \end{subfigure}
     \caption{Visual exploration of various attribute distributions present within \dataset.}
     \label{fig:datastats}
\end{figure}

Figure \ref{fig:datastats} gives an overview of our dataset: \dataset. Figures \ref{fig:datastats1} and \ref{fig:datastats2} show the distributions of the target communities in the hate speech and intents across the counterspeeches, respectively.

For a more fine-grained perspective, Figure \ref{fig:datastats3} and \ref{fig:datastats4} show the uniform distributions of intents in the data splits and the intents across target communities. Figures \ref{fig:datastats5} and \ref{fig:datastats6} depicts the average token lengths for the five intent classes and eight target communities.


\section{Additional Details on Experiments}\label{app:detail}
\paragraph{Experimental Setup:}
All the experiments were performed using a Tesla V100 and an RTX A6000 GPU. Our model (and the BART baseline) was trained for 20 epochs with the initial learning rate of 8e-5 using AdamW as the optimizer and a linear scheduler, with 10\% of the total steps as warm-up having a weight decay of 0.03. Training the model took an average time of 3 hours with a batch size of 32, and the model with the best validation loss was employed for testing. We used the base version of BART (140M parameters) from the transformers library \cite{wolf-etal-2020-transformers} for parameterizing both $\phi_s$ and $\phi_i$. The baselines were trained using the recommended hyperparameter settings. To compute the ROUGE score, we use the rouge library in python with the default arguments, we compute METEOR through nltk \cite{bird2009natural}, semantic similarity by using the \textit{all-miniLM-v2} model from the sentence-transformers library \cite{reimers-gurevych-2019-sentence} and BERTScore using the original bert-score library. To check the efficacy of the models in incorporating the desired intent in the generated counterspeeches, we train an Intent Classification (IC) model on \dataset for intent classification of each counterspeech instance, which achieves 75\% accuracy on the test set for classification (we utilize the base version of RoBERTa). The IC model is used to classify whether the generated counterspeeches are compatible with the desired intent, and the accuracy obtained across the generated samples is reported as the \textit{category accuracy}.

\begin{figure*}[h]
     \centering
          \begin{subfigure}[b]{0.3\textwidth}
         \centering
         \includegraphics[width=\textwidth]{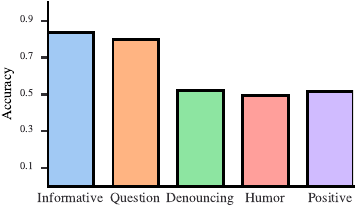}
        \caption{A fine-grained analysis of intent identification accuracy of the generated outputs from \model on the test set as per the $IC$ model.}
        \label{fig:ca-bar}
     \end{subfigure}\hspace{1em} 
     \begin{subfigure}[b]{0.3\textwidth}
         \centering
         \includegraphics[width=\textwidth]{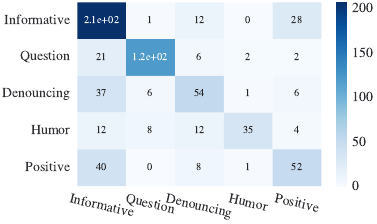}
        \caption{Confusion matrix depicting the intent classification (from the $IC$ model) of the generated outputs from \model.}
        \label{fig:cm-ca}
     \end{subfigure}
     \caption{Automated evaluation of $CA$ from the $IC$ model for all intents. Note that \textit{informative} and \textit{question} achieve the highest accuracy demonstrating that \model is able to generate them more effectively than, say, humor, which achieves a relatively lower accuracy.}
     \label{fig:automated-ca}
\end{figure*}

\begin{figure*}[h]
     \centering
     \begin{subfigure}[b]{0.3\textwidth}
         \centering
        \includegraphics[width=\columnwidth]{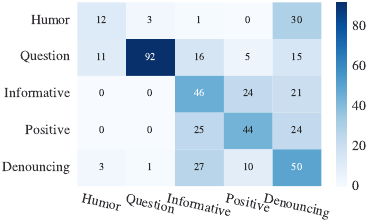}
        \caption{Confusion matrix of the human evaluation for \model.}
        \label{fig:cm-quar-human}
     \end{subfigure}\hspace{2em}
   \begin{subfigure}[b]{0.3\textwidth}
 \centering
    \includegraphics[width=\columnwidth]{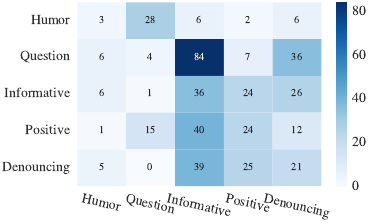}
    \caption{Confusion matrix of the human evaluation for the best baseline, GPS.}
    \label{fig:cm-gps-human}
     \end{subfigure}
     \caption{Human evaluation heatmaps for \model and GPS. The rows represent the desired intent (the input given to the models) and the columns denote the intent labeled by human evaluators. Darker shade denotes a higher frequency of identification. For \model, all intents but \textit{humor} are generally identifiable, while GPS is unable to condition on any intent effectively.}
     \label{fig:human-eval}
\end{figure*}

\section{Analysis of Intent-Conditioning}
\label{appendix:analysisIC}
In order to systematically evaluate the effects of intent conditioning, we begin by analyzing the accuracy of the  $IC$ model for each intent separately. The results are depicted in Figure \ref{fig:automated-ca}. From the bar chart, we observe that the accuracy of the intents -- \textit{informative} and \textit{question}, is higher than the other intents, while \textit{humor} displays the lowest accuracy. To obtain a more comprehensive understanding, the confusion matrix illustrates that the intents \textit{denouncing} and \textit{positive} tend to be recognized as \textit{informative} by the $IC$ model in some cases, while \textit{humor} can also be recognized as \textit{informative} and \textit{denouncing}. Since the $IC$ model is susceptible to errors, it is hard to say with certainty whether the generated counterspeech belongs to the desired intent, or whether the model has misclassified it. Hence, we utilize the confusion matrices from human evaluation and design a new metric in the next section for analyzing the intent conditioning due to the inherent reliability of human evaluators.

\begin{figure*}[h]
     \centering
     \begin{subfigure}[b]{0.3\textwidth}
         \centering
         \includegraphics[width=\textwidth]{analysis_plots/codebook_embedding.pdf}
        \caption{A scatter plot of the codebook vectors corresponding to different intents after being reduced to a two-dimensional space through Principal Component Analysis (PCA).}
        \label{fig:codebook-embed}
     \end{subfigure}\hspace{1em}
      \begin{subfigure}[b]{0.3\textwidth}
 \centering
    \includegraphics[width=\textwidth]{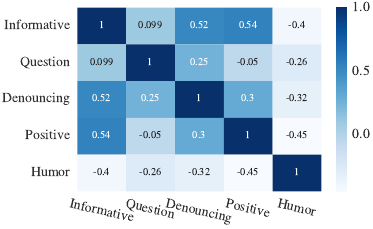}
    \caption{Visualization of the cosine similarity between the codebook vectors corresponding to different intents. Darker shade denotes higher similarity.}
    \label{fig:cos-sim}
     \end{subfigure}\hspace{1em}
\begin{subfigure}[b]{0.3\textwidth}
 \centering
    \includegraphics[width=\textwidth]{analysis_plots/impl_sim_full.pdf}
    \caption{The captured Implicit Similarity between all pair of intents. Note that $(I,D)$ and $(I,P)$ achieve the highest scores, while $(P,H), (Q,P), (I,H)$ and $(Q,H)$ achieve the lowest scores.}
    \label{fig:cos-sim}
     \end{subfigure}
     \caption{Analysis and visualization of intent representations through: (a) dimensionality reduction to a 2-D space for plotting; (b) cosine similarity computed in the original dimension space of the representations. The similarity between \textit{informative}, \textit{positive} and \textit{denouncing} is higher as compared to other intents. (c) The $IS$ scores are closely aligned with the closeness of the representations in (a) and cosine similarities in (b). This serves to inform that the quantized representations learnt for each intent are demonstrably sound due to their similarity with human feedback.}
     \label{fig:codebook-sim}
\end{figure*}

\section{Interpretability and Robustness of Intent Representations}
\label{appendix:is}
A key advantage afforded by our approach is the exploration of interpretability, which is enabled by our paradigm of learning the intent representations separately. The intent representations illustrated in Figure \ref{fig:interpretability} (\textit{left}) depict that the intents \textit{positive} and \textit{denouncing} are both mapped closely to \textit{informative}, and are slightly farther away from each other, while \textit{question} and \textit{humor} are considerably distant to all other intents. This observation is further supported by computing the cosine similarity in the original dimension of the representations (Fig. \ref{fig:cos-sim}). To assess the robustness of the obtained representations, we use implicit feedback from human evaluations to gauge the similarity between intents. We employ two strategies: (i) we design a new metric, \textit{Implicit Similarity (IS)} to compute the similarity between pairs of intents implicitly through human evaluation responses without the knowledge of the actual intent; (ii) we utilize the intent information and use the confusion matrices obtained from human evaluation (Fig \ref{fig:cm-quar-human}) for this purpose.

We plot the $IS$ values for each intent pair in Figure \ref{fig:codebook-sim}. The $IS$ scores for the pairs $(I, D)$ and $(I, P)$ are the highest, followed by the pair $(P,D)$, while the lowest scores are achieved by the pairs $(P,H), (Q,P), (I,H)$ and $(Q,H)$. Interestingly, the $IS$ scores \textit{closely align} with the distances between the intent representations in the scatter-plot in Figure \ref{fig:interpretability}. This demonstrates the robustness of the intent representations learned by \model and highlights a critical factor responsible for its performance, as the representations align with the proximity that is inherently captured by evaluators.

\paragraph{Explicit Similarity through Human Evaluation:} To further analyze the intent representations, we also utilize the desired intent $c_i$ to generate the confusion matrices for human evaluation in Figure \ref{fig:human-eval}. We observe a similar pattern to that observed through $IS$, as we can see that the bottom-right $3\times 3$ square has a darker shade as compared to the rest of the matrix, denoting that the \textit{Informative, Positive} and \textit{denouncing} intents are closer together when compared to other pairings.

\section{Human Evaluation}
\label{appendix:humanEval}
The evaluators recruited were well-versed in the field of NLP and social media. The form provided to them contained the descriptions of terminology such as \textit{Hate Speech} and \textit{Counterspeech}, and \textit{Intents}. For further clarity, a few lines of description for each intent along with an example were also shown. The form also included information on the format of the questionnaire; the evaluators were  made aware of how the evaluation data would be used in the study and were warned against the possibility of encountering foul or offensive language that could be upsetting.

\paragraph{Analysis:} As shown in Figure \ref{fig:human-eval}, our model generates intent-identifiable outputs across all intents, with the exception of the \textit{humor}, where the outputs were often assigned to \textit{denouncing}. Conversely, GPS fails to effectively condition on intent, as evidenced by the mismatch between desired and obtained intents, with decent performance only on \textit{informative}, perhaps due to its prevalence in the training set.


\end{document}